\documentclass[letterpaper, 10 pt, conference]{ieeeconf}  

\IEEEoverridecommandlockouts                              

\usepackage{amsmath} 
\usepackage{graphicx}
\usepackage{color}
\usepackage{booktabs}
\usepackage{array}
\usepackage[table]{xcolor}
\usepackage{amssymb}
\usepackage{multirow}
\usepackage{booktabs}
\usepackage{caption}
\usepackage{subfig}
\usepackage[none]{hyphenat}

\title{\LARGE \bf
Bi-Hap: a Bi-directional Momentum-based Haptic Feedback and Control System for Dexterous In-hand Telemanipulation}

\author{Haoyang Wang$^{1}$, Haoran Guo$^{1}$, He Bai$^{1}$, \textit{Member, IEEE}, Zhengxiong Li$^{2}$, \textit{Member, IEEE},\\and Lingfeng Tao$^{1}$, \textit{Member, IEEE}
\thanks{*This work is supported by NSF \#2426269 and \#2426470.}
\thanks{$^{1}$H. Wang, H. Bai and L. Tao are with the Oklahoma State University, 563 
Engineering North, Stillwater, OK, 74078, USA (e-mail: haoyang.wang;
haoran.guo; he.bai; lingfeng.tao@okstate.edu).
}%
\thanks{$^{2}$Z. Li is with the University of Colorado Denver, I Department of Computer Science and Engineering, 1380 Lawrence St. Center, LW-834, Denver. CO 80217, USA (e-mail: zhengxiong.li@ucdenver.edu).
}}%

\begin{document}

\maketitle
\thispagestyle{empty}
\pagestyle{empty}


\noindent \begin{abstract}
In-hand dexterous telemanipulation requires not only precise remote motion control of the robot but also effective haptic feedback to the human operator to ensure stable and intuitive interactions between them. Most existing haptic devices for dexterous telemanipulation focus on force feedback and lack effective torque rendering, which is essential for tasks involving object rotation. While some torque feedback solutions in virtual reality applications—such as those based on geared motors or mechanically coupled actuators—have been explored, they often rely on bulky mechanical designs, limiting their use in portable or in-hand applications. In this paper, we propose a Bi-directional Momentum-based Haptic Feedback and Control (Bi-Hap) system that utilizes a palm-sized momentum-actuated mechanism to enable real-time haptic and torque feedback. 
The Bi-Hap system also integrates an Inertial Measurement Unit (IMU) to extract the human's manipulation command to establish a closed-loop learning-based telemanipulation framework. 
Furthermore, an error-adaptive feedback strategy is introduced to enhance operator perception and task performance in different error categories. Experimental evaluations demonstrate that Bi-Hap achieved feedback capability with low command following latency (Delay $<$ 0.025 s) and highly accurate torque feedback (RMSE $<$ 0.010 Nm).

\end{abstract}


\section{INTRODUCTION}

\noindent Dexterous in-hand telemanipulation has emerged as a key capability in domains such as human-robot collaboration \cite{huang2025human}, advanced telerehabilitation \cite{cortese2014mechatronic}, and imitation learning data collection \cite{wang2024dexcap}, where precise and intuitive control of an object’s pose within the hand is essential. Unlike conventional teleoperation tasks using simple grippers or tools, in-hand manipulation requires dexterous skills such as rotation, re-grasping, and fine adjustment. Achieving this remotely demands precise motion control and realistic haptic feedback—especially torque feedback—to help the operator perceive subtle rotational dynamics and maintain stable, accurate control. 
However, existing haptic devices mainly focus on force feedback and lack torque rendering, which is critical for object rotation. For example, DoGlove \cite{zhang2025doglove} provides multi-finger force sensing but no torque feedback, while DextrES \cite{abad2022novel} simulates grasping force via electrostatic brakes without rotational cues, hindering the operator’s ability to perceive and manipulate rotational dynamics during complex telemanipulation tasks.
Some existing prototypes attempt to provide torque feedback but rely on large, complex mechanical structures \cite{liu2014development} that are unsuitable for in-hand applications. These limitations create a significant gap between the manipulation dexterity and the feedback capabilities of current telemanipulation systems, impeding the operator’s ability to perform fine manipulation tasks.

\captionsetup{font=footnotesize}
\begin{figure}[tp]
    \centering
    \vspace{7pt} 
    \includegraphics[width=8cm]{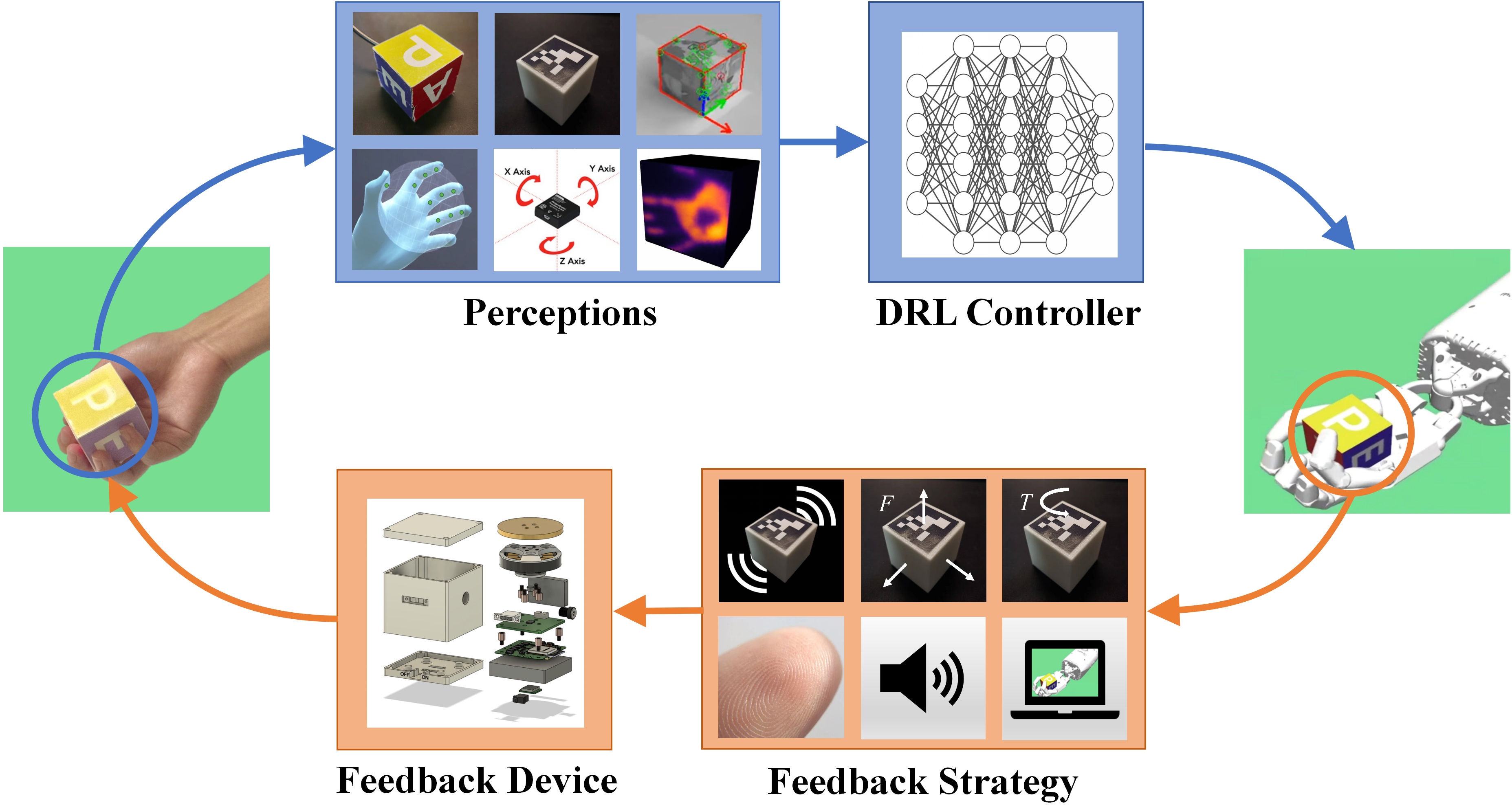}
    \vspace{-3pt}
    \caption{The data flow in the proposed Bi-Hap system establishes a bi-directional feedback and control loop, enabling not only DRL-based teleoperation from the human to the robot but also recreating a seamless manipulation experience through torque and vibration feedback delivered from the robot back to the human.}
    \vspace{-0.7cm} 
    \label{fig:sora_system}
\end{figure}

Momentum-based haptic devices have shown great potential in the domain of torque feedback, offering an innovative approach to simulating torque and vibrations in interactive systems \cite{meijneke2021design}. 
Unlike conventional haptic devices that depend on motors or actuators fixed to the ground for direct force application, momentum-based systems generate inertial forces using spinning masses like momentum wheels. This allows for a compact and portable design while delivering rich haptic feedback, including torque, impact simulation, and both low-frequency and high-amplitude vibrations.
Despite these advantages, momentum-based haptic devices have not yet been explored in the context of dexterous in-hand telemanipulation. One reason is that most existing devices require full-palm grasping \cite{berna2024hapticwhirl, hashimoto2022metamorphx}, and their irregular shapes make it difficult for fingers to manipulate them dexterously within the hand. Additionally, the complexity of controlling such systems due to the inherent inertia of the spinning masses leads to feedback latency, which is critical in dynamic telemanipulation tasks.

To bridge these gaps, we propose a \textbf{Bi-directional Momentum-based Haptic Feedback and Control (Bi-Hap)} system for real-time, in-hand dexterous telemanipulation. The Bi-Hap system leverages a lightweight (less than 200g), palm-sized momentum-actuated mechanism to generate perceivable torque feedback directly to the operator’s hand without relying on external grounding. 
Additionally, an inertial measurement unit (IMU) is integrated to capture human-object interaction and transmit it to a deep reinforcement learning (DRL) controller, enabling seamless and intuitive telemanipulation \cite{wang2024real}.
Furthermore, we developed an error-adaptive feedback strategy for the Bi-Hap system, dynamically adjusting the feedback based on the manipulation error states to enhance the operator’s perception and decision-making. 
Through telemanipulation experiments involving human operators, we demonstrate that the proposed system enhances both the precision and situational awareness of the operator in in-hand object rotation tasks.

The contributions of this work are summarized as follows:

1). We designed a palm-sized Bi-Hap system for dexterous in-hand telemanipulation, capable of rendering real-time torque and vibration feedback without external grounding.

2). We developed an error-adaptive feedback strategy that dynamically modulates feedback based on manipulation error states to improve operator perception and task performance.

3). We validated the effectiveness of the Bi-Hap system through comprehensive experiments in real-world scenarios, demonstrating improved manipulation accuracy, responsiveness, and task performance.

\section{RELATED WORK}

\subsection{Telemanipulation with Haptic Feedback}

\noindent Haptic feedback has long been recognized as a critical component to enhance telemanipulation performance \cite{pacchierotti2023cutaneous}, particularly in tasks demanding fine dexterity and interaction with objects of complex geometries. 
By providing the human operator with real-time tactile and kinesthetic cues, haptic systems improve the perception of contact forces, object stiffness, and dynamic interaction, thereby facilitating more intuitive and stable control \cite{zhang2025doglove}. 

Prior studies have shown that the inclusion of force feedback in teleoperation loops significantly reduces task completion time and error rates, especially in delicate manipulation scenarios such as needle insertion \cite{raitor2023design}, object alignment \cite{quek2018evaluation}, and vehicle assembly \cite{sagardia2016platform}. 
Traditional haptic interfaces—such as grounded force-feedback arms or exoskeleton gloves—have demonstrated the benefits of bidirectional information flow. However, there remains a significant gap in providing effective torque feedback, which is crucial for tasks involving object rotation within the hand—such as turning a screwdriver, twisting a bottle cap, or reorienting an irregular-shaped object \cite{krieger2024motion}. While some high-DOF haptic gloves attempt to emulate torque sensations by applying 3D fingertip forces \cite{liu2014development}, this indirect approach often lacks realism and fidelity. Moreover, such systems typically rely on bulky, grounded actuation setups, limiting their practicality for portable or in-hand telemanipulation applications where lightweight and integrated feedback solutions are essential.

\subsection{Momentum-based Haptic Feedback Device}

\noindent Momentum-based haptic devices offer a promising solution for ungrounded torque rendering. By leveraging rotating flywheels or control moment gyroscopes (CMGs), these systems can generate perceptible torques without requiring a physical ground \cite{meijneke2021design}, making them suitable for in-hand or mobile contexts.

Several studies have demonstrated the feasibility of such designs for mobile and handheld applications. For example, Walker et al. \cite{walker2017haptic} developed a compact dual-CMG haptic device that delivers clear directional torque cues by rapidly reorienting counter-rotating flywheels through asymmetric pulses. Similarly, Berna Moya et al. \cite{berna2024hapticwhirl} proposed HapticWhirl, a handheld controller that uses a flywheel and dual-axis gimbal to deliver multimodal haptic feedback including torque, vibration, and inertial cues. Notably, Hashimoto et al. \cite{hashimoto2022metamorphx} designed MetamorphX, an ungrounded 3-DoF moment display utilizing four control moment gyroscopes (CMGs) to enable low-latency rotational impedance feedback, offering realistic inertia and viscosity changes suitable for VR object manipulation.

These designs offer promising avenues for wearable or handheld torque feedback in virtual reality (VR) systems. However, they are not specifically tailored for dexterous in-hand telemanipulation. The devices are typically grasped by the whole hand \cite{berna2024hapticwhirl, hashimoto2022metamorphx, verret2024synthesis}, with torque cues primarily perceived through wrist rotation rather than finger-level interaction. Moreover, their overly heavy mechanical structures and complex geometric shapes make it difficult to perform fine in-hand movements such as translation and reorientation using the fingers alone. This limits their applicability in scenarios requiring precise object manipulation within the operator’s palm.

\section{BI-DIRECTIONAL HAPTIC FEEDBACK AND CONTROL SYSTEM}
\vspace{-3pt}
\subsection{Hardware Implementation}
\noindent The Bi-Hap system features a modular design that supports various motors and sensors for different tasks and can be scaled to more degrees of freedom by replacing the current motor with smaller, higher-speed servo motors. Fig.~2 presents the system architecture in block diagram form. The main modules are as follows:

\textbf{Host Computer Module:} The host computer serves as the central command unit, which receives signals from sensors such as the IMU and controls the robot via the DRL controller \cite{wang2024real}. It also transmits feedback intensity and adjusts control gains through UDP and TTL communication. UDP ensures low-latency, wireless data transmission, while TTL facilitates wired program uploading and debugging. This dual-channel setup enhances development flexibility and communication reliability.

\textbf{Control Module:} An ESP32 microcontroller manages the motor drive circuit, executing the closed-loop Field Oriented Control (FOC) algorithm using the SimpleFOC library \cite{skuric2022simplefoc}. This allows for stable and precise torque control of the BLDC gimbal motor, especially at dynamic target velocities. The ESP32 also handles real-time data from sensors and executes control logic with minimal latency, benefiting from its integrated Wi-Fi and SPI/I2C support.

\textbf{Torque Generating Module:} The actuator consists of a flat gimbal brushless DC (BLDC) motor known for its compact size and high torque density, coupled with a 55 mm flywheel disk as the counterweight. The flywheel is chosen to balance overall weight (128 g) and rotational inertia ($4.8 \times 10^{-5}$ kg·m²), enabling effective momentum-based haptic feedback. A 14-bit AS5047 magnetic encoder provides precise angular feedback via SPI.

\textbf{Power Management Module:}
This module includes a 12V battery and a protection board, supplying 12V to the BLDC motor and converting it to 5V for control and sensor electronics. A power switch and DC interface allow convenient charging or direct power input, ensuring safe and stable energy distribution during operation.

\textbf{Sensor Module:} The module currently integrates an IMU for 6-DoF attitude sensing at 200 Hz via I2C, supporting real-time motion tracking and closed-loop control. It also allows further expansion through SPI or I2C interfaces, enabling the integration of additional sensors, such as temperature or pressure sensors, for customized perception.

\captionsetup{font=footnotesize}
\begin{figure}[t]
    \centering
    \vspace{0pt} 
    \includegraphics[width=7cm]{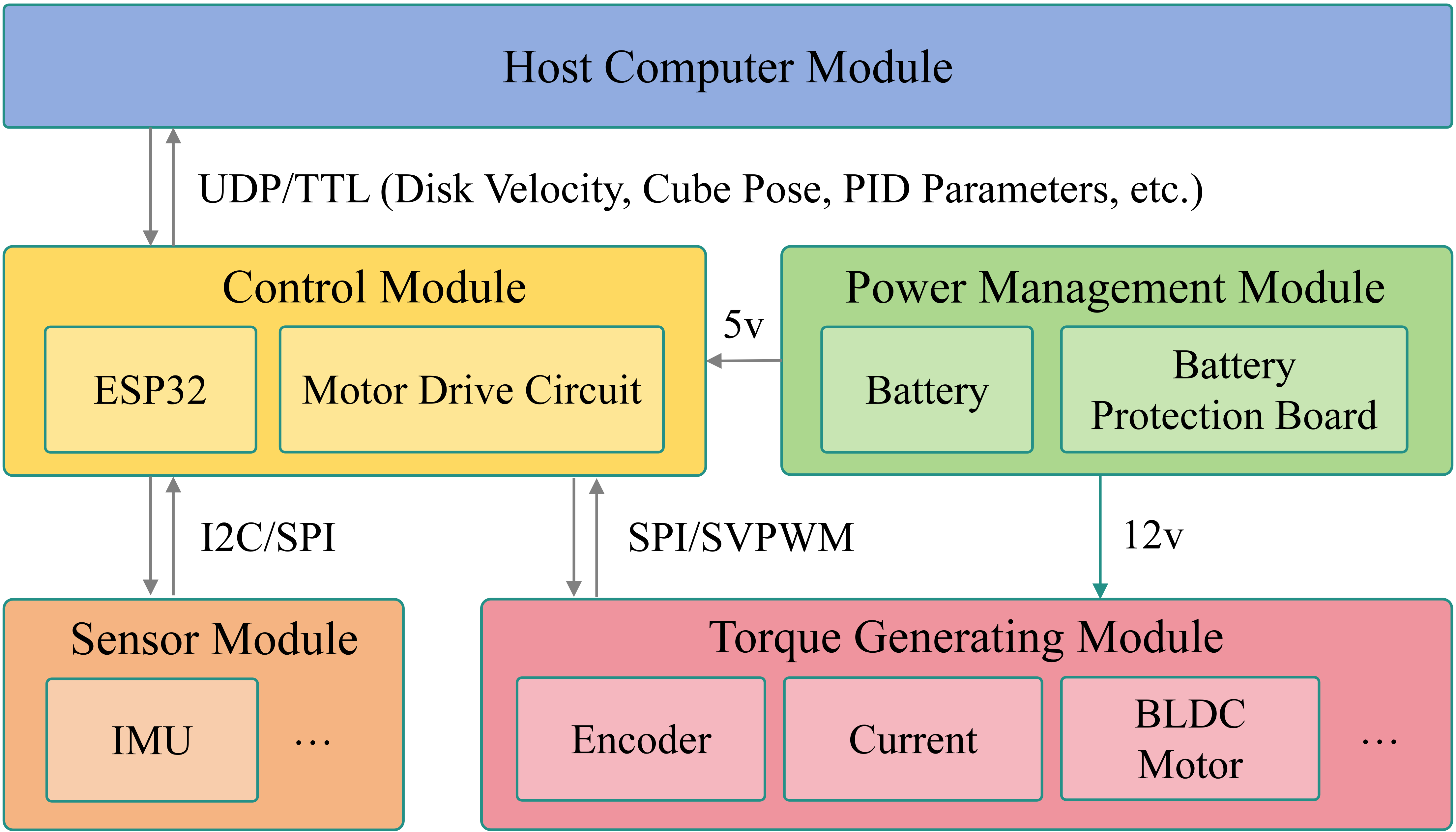}
    \caption{System architecture of the Bi-Hap system, illustrating key modules for control, sensing, power, and torque feedback.}
    \vspace{-0.7cm} 
    \label{fig:sora_system}
\end{figure}

An exploded view of the Bi-Hap system is shown in Fig. 3. Including the counterweight, the total weight of the Bi-Hap system is 320 g. To protect the internal components, facilitate intuitive finger manipulation by human operators, and enable convenient arrangement and replacement of internal parts, we designed a three-layer cubic structure with a side length of 60 mm, consisting of upper, middle, and lower sections. The upper layer can be removed to replace the counterweight, while the lower layer provides access for battery and sensor replacement. This modular design also allows for easy disassembly and maintenance, enhancing the system’s usability and serviceability.

\subsection{Working Principle}
\noindent According to the law of conservation of angular momentum, in the absence of external torque, a system’s total angular momentum remains constant. When a motor accelerates a flywheel, it gains angular momentum, which is balanced by equal and opposite angular momentum in the motor or its support structure. The torque resulting from this change in angular momentum is calculated as:
\begin{equation}
\tau = -I_f \cdot \frac{d\omega_f}{dt}
\end{equation}
where $\tau$ represents the counteracting torque on its supporting structure, $I_f$ is the moment of inertia of the flywheel, and $\frac{d\omega_f}{dt}$ is the rate of change of the flywheel's angular velocity.

To regulate torque, we adopt velocity control, as torque is linked to angular acceleration. A discrete Proportional Integral Derivative (PID) controller adjusts motor voltage based on the velocity error, with control input at step $k$ defined as:
\begin{equation}
u(k) = K_p e(k) + K_i \sum_{i=0}^{k} e(i) \Delta t + K_d \frac{e(k) - e(k-1)}{\Delta t}
\end{equation}
At each discrete time step $k$, the control output $u(k)$ represents the motor voltage, calculated based on the error $e(k)$ between the desired and actual flywheel angular velocities. A discrete PID controller is employed, with gains $K_p = 0.2$, $K_i = 20$, and $K_d = 0$. The controller is initially tuned using the Ziegler–Nichols method \cite{patel2020ziegler}, then refined manually to accommodate system-specific inertia and damping. The proportional term ensures rapid response, the integral term eliminates steady-state error, and the derivative term was found unnecessary for stability. This control strategy effectively regulates angular velocity to produce the desired feedback torque for the haptic interface.

\captionsetup{font=footnotesize}
\begin{figure}[t]
    \centering
    \vspace{0pt} 
    \includegraphics[width=7cm]{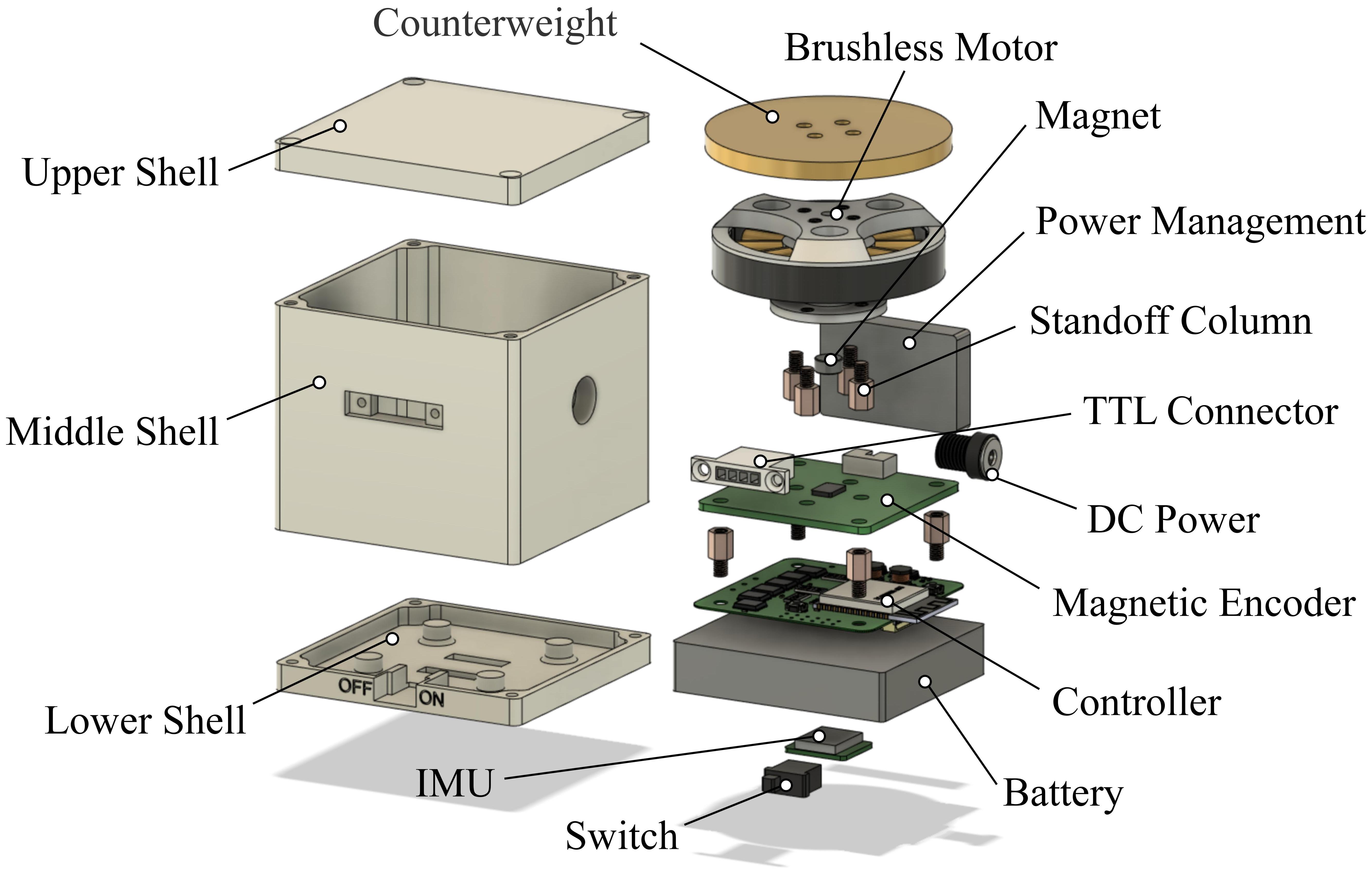}
    \vspace{-2pt}
    \caption{Exploded view of the Bi-Hap device with three-layer cubic structure (side length = 60 mm), for easy maintenance, component replacement, and intuitive in-hand telemanipulation.}
    \vspace{-0.6cm} 
    \label{fig:sora_system}
\end{figure}

To enhance the intuitiveness of interaction between the Bi-Hap system and the human operator, an improved impedance control strategy is employed to compute the feedback torque. This method simulates a spring-damper system, adapting the torque response to external disturbances during task execution. The desired torque is calculated as:
\begin{equation} 
\tau_d = K_{rot}(\theta_d - \theta) + B_{rot}(\dot{\theta_d} - \dot{\theta}) + M_{rot}(\ddot{\theta_d} - \ddot{\theta}) 
\end{equation}
where $\tau_d$ denotes the desired torque; $\theta$, $\dot{\theta}$, and $\ddot{\theta}$ represent the current angular position, velocity, and acceleration of the in-hand object, respectively; and $\theta_d$, $\dot{\theta_d}$, and $\ddot{\theta_d}$ are their corresponding desired values. The coefficients $K_{rot}$, $B_{rot}$, and $M_{rot}$ represent the rotational stiffness, damping, and inertia, respectively
The control gains were calibrated through experiment to ensure both responsiveness and stability, with $K_{\mathrm{rot}} = 2.5$, $B_{\mathrm{rot}} = 1$, and $M_{\mathrm{rot}} = 0$.

To prevent flywheel speed saturation, the target torque adjusts the output mode rather than being directly applied. When the flywheel is unsaturated, torque is applied normally. If saturated, the torque magnitude is converted into a vibration amplitude for output, which is calculated as:
\begin{equation}
\tau_d = \alpha\sin(\omega t)
\end{equation}
where $\tau_d$ represents the final output signal, which is the torque feedback provided to the operators. $\alpha$ is the maximum amplitude of the vibration, $\omega$ is the angular frequency of the vibration, and $t$ is the time variable.

\subsection{Error-Adaptive Feedback Strategy}

\noindent In our telemanipulation setup, human operators use the Bi-Hap system to remotely rotate the object in the robot hand to the target angle.
To ensure precision and reduce operator burden, we introduce an error-adaptive feedback strategy that adjusts the modality and intensity of feedback based on the current manipulation state. 
By integrating visual, auditory, and torque cues according to the orientation error, we identify the following manipulation scenarios and design the corresponding feedback strategies:

\textbf{Target Position Reached:}
Auditory and visual feedback are provided in this scenario ($error < 1.6^{\circ}$).
Auditory feedback includes a short, sharp ``ding" sound to notify the human operator that the target angle is reached.
In this case, torque feedback is unnecessary as it may cause the manipulated object to deviate from the target angle.

\textbf{Target Position Overshot:}
Torque and visual feedback are provided in this scenario.
The Bi-Hap system generates torque, with magnitude calculated by Equation (3), in the direction of the target to prompt the human operator to make the necessary adjustments.

\textbf{Normal Manipulation:} 
Only visual feedback is provided in this scenario ($1.6^{\circ} < error < 3.2^{\circ}$), which suits cases where the object is near the target angle.
Visual cues alone can convey the angle difference to the operator, while excessive feedback may increase cognitive load unnecessarily.

\textbf{Far from Target Position:}
Torque and visual feedback are provided in this scenario ($error > 3.2^{\circ}$).
If the error changes rapidly and the operator continues rotating away from the target, the Bi-Hap device applies torque—magnitude from Equation (3)—toward the target to signal a major correction. Feedback activates only if the operator remains in the deviation area for over 3 seconds, based on user input, to avoid overwhelming them with frequent corrections.

\textbf{Failure Movements:}
Visual, auditory, and vibration feedback are provided in this scenario. Due to current technological limitations, it is difficult for the robotic hand to match the dexterity of a human hand, which can lead to unsafe rotations or displacements of the object. When the object's position moves outside a specific range on the robot's hand—either due to it being stuck or falling off—this type of feedback is provided to alert the operator to take corrective action.

\section{Experiments Design and Evaluation Metrics}

\begin{figure}[t]
    \vspace{-5pt}
    \centering
    \subfloat[Human offline test]{\includegraphics[width=0.187\textwidth]{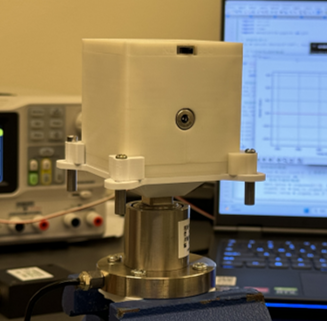}}
    \hfill
    \subfloat[Human online test]{\includegraphics[width=0.27\textwidth]{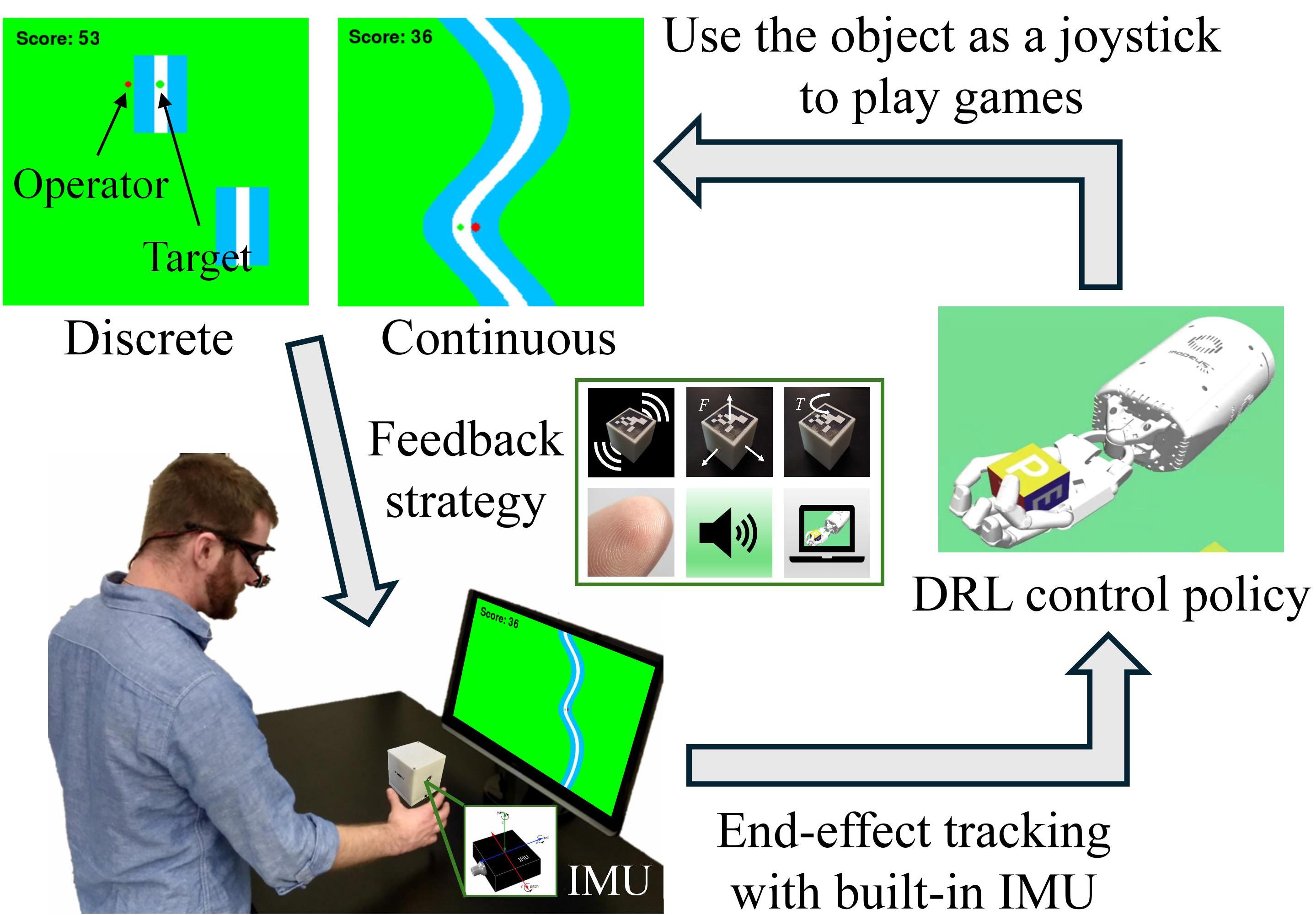}}
    \vspace{-1pt}
    \caption{Experimental setup for: (a) human offline test; (b) human online test. The white area in Fig. 4(b) represents the target position, and points are awarded when the red dot (represents the operator's current position) enters this area, corresponding to \textit{Target Position Reached}. Exceeding the boundary of the white area corresponds to \textit{Target Position Overshot}. The blue area serves as a warning zone, reminding the operator that adjustments are needed, corresponding to \textit{Normal Manipulation}. The green area represents the forbidden zone, corresponding to \textit{Far From Target Position}.}
    \label{fig:teleplay}
    \vspace{-15pt}
\end{figure}

\subsection{Experimental Setup and Task Design}
\noindent To comprehensively evaluate the performance of the Bi-Hap device, we divide the testing into two categories:

\textbf{Human Offline tests:} The human offline tests refer to those that do not require the involvement of a human operator (like Fig. 4(a)). These include measurements of the accuracy of torque feedback, the speed at which torque is generated, and the control system's latency. The testing platform incorporates a ±2 Nm torque sensor and a 3D-printed mounting structure. With a 500 Hz sampling rate, the sensor captures signal frequencies up to 250 Hz, as dictated by the sampling theorem. This range covers the low-frequency (0-10 Hz) and mid-frequency (10-100 Hz) forces, which are most perceptible to human touch \cite{bai2023robotic}. The torque sensor's accuracy is 0.1\%, which is below the 1\% force variation threshold that human touch can detect \cite{wang2024toward}. The Bi-Hap device is configured to follow the following test signals:

\subsubsection{Sinusoidal goal} The torque generated by the feedback device is set to follow the following sinusoidal trajectory:
\vspace{-2pt}
\begin{equation}
\tau_d = \alpha \sin(\omega x)
\end{equation}
where $\alpha = (0.015,\ 0.030)$ is the amplitude, and $\omega = (8,\ 16)$ is the frequency. The sinusoidal goal helps measure performance under periodic torque feedback.

\subsubsection{Square wave goal} The torque generated by the feedback device is set to follow the following square wave trajectory:

\begin{equation}
\tau_d = \left\{ \begin{aligned}
A,\ \  & \text{if } ((5t \bmod 2\pi) < \pi) \\
-A,\ \  & \text{if } ((5t \bmod 2\pi) \geq \pi)
\end{aligned} \right.
\end{equation}
where $A = (0.01,\ 0.02)$, $t$ is the time variable. The square wave goal helps to evaluate the feedback performance for targets that change in a short time.

\textbf{Human Online Tests:} 
This experiment aims to evaluate the performance of the feedback device during real human interaction (like Fig. 4(b)). 
We designed discrete and continuous scenarios to represent the two most common cases in dexterous telemanipulation. 
After completing each task, a score is given to indicate the overall teleoperation performance. 
As an ablation study, the operator will also use a version of the Bi-Hap device with the same size and weight but with torque feedback disabled to control the robotic hand in target-following tasks. Finally, the operator's performance in these two conditions will be analyzed.
The robotic control tasks are performed in simulation using the Shadow Hand environment on the MuJoCo physics simulator \cite{todorov2012physics}.
The specific task is conducted in a gaming environment, where the human operator holds the Bi-Hap device as a joystick, and a DRL controller drives the robotic hand to rotate the cube to match the orientation of the Bi-Hap device. Meanwhile, the cube's angle in the robotic hand is mapped to the movement of a red dot on the screen. The operator must rotate the Bi-Hap device to guide the robotic hand in aligning the object with the target orientation.
Following the university IRB protocol, five participants with prior telemanipulation experience were recruited. Each was thoroughly familiarized with the setup before conducting five trials under four conditions: discrete vs. continuous tasks, with and without torque feedback.

\vspace{0pt}
\subsection{Evaluation Metrics}
\noindent The following metrics are designed for the sinusoidal goal:

\noindent\underline{\textit{RMSE:}} The root mean square error is calculated as:
\begin{equation}
RMSE = \sqrt{\frac{1}{n} \sum_{i=1}^{n} (\tau_{a,i} - \tau_{d,i})^2}
\end{equation}
where \( n \) is the number of timesteps, $\tau_{a,i}$ is the measured torque around the Z-axis at the $i$-th sampling, and $\tau_{d,i}$ is the desire torque at the $i$-th sampling. 
RMSE evaluates the overall torque feedback accuracy during the testing process.

\noindent\underline{\textit{Average Latency:}} The average latency is calculated as:
\begin{equation}
L = \frac{1}{n} \sum_{i=1}^{n} \frac{\varphi_{a,i} - \varphi_{d,i}}{2\pi f}
\end{equation}
where $\varphi_{a,i}$ is the phase of the actual torque variation trajectory at the $i$-th sampling, $\varphi_{d,i}$ is the phase of the desired torque variation trajectory at the $i$-th sampling. Both phases are obtained from the dominant frequency after the Fourier transformation. The average latency evaluates the manipulation delay, which is critical for real-time capability.

The following metrics are designed for the square wave:

\noindent\underline{\textit{RMSE:}} This metric is calculated the same as the sinusoidal.

\noindent\underline{\textit{Overshoot:}}  We calculate the max-percent overshoot:

\begin{equation}
OS = \frac{\tau_a - \tau_d}{\tau_d} \times 100\%
\end{equation}

where $\tau_a$ denotes the actual peak torque, and $\tau_d$ is the desired torque amplitude, equal to the square wave amplitude $A$. This metric reflects the aggressiveness of the Bi-Hap system in response to abrupt torque changes.

\noindent\underline{\textit{Peak Time:}} The time $t_p$ taken for the response to reach its first peak value. It reflects how rapidly Bi-Hap reacts to torque feedback commands. For all the above metrics, lower values indicate better performance.

The following metrics are designed for human online tests:

\noindent\underline{\textit{Fit Count:}} We record the number of times the target point is moved into the designated area in each experiment as a measure of overall performance. The total number of target completions is denoted as $FC:$
\vspace{-3pt}
\begin{equation}
FC = \sum_{i=1}^n S_i
\end{equation}
where $n$ represents the total number of timesteps. The target angle must be reached within a $\pm 1.6^{\circ}$ range and held for 1.6 seconds. During each refresh, the system checks if the object's angle falls within the target range. If it does, $S_i = 1$; if not, $S_i = 0$, where $i$ represent the refresh count.

\section{Results and Discussion}
\subsection{Human-Offline Testing}
\noindent The quantitative results for sinusoidal goal tracking are shown in Table~\ref{tab:sinusoidal_tracking}. Bi-Hap achieves low latency (max 0.025 s) and high accuracy (max 0.0095 Nm), demonstrating fast and precise torque feedback. More visual results are provided in the supplementary video. In comparison, systems like MetamorphX~\cite{hashimoto2022metamorphx} (delay $\approx 0.060$ s) would require a threefold increase in response speed to match Bi-Hap’s performance.
The tracking performance deteriorates as the sinusoidal amplitude increases and improves as the sinusoidal frequency decreases, owing to the alteration in task complexity.

As shown in Table~\ref{tab:step_tracking_5tests}, compared to the sinusoidal tracking results, the higher overshoot (OS) during step response tests mainly arises from the motor’s torque limitations, which hinder rapid responses to abrupt changes. Nevertheless, the system exhibits consistent behavior across different step amplitudes, indicating stability under varying demands. Notably, larger step inputs yield a lower OS ratio, demonstrating robustness in tracking larger changes. The variation in peak time $t_p$ reflects the expected trade-off between target magnitude and response speed. While RMSE increases with amplitude due to higher torque demands, overall tracking remains within acceptable bounds.

\begin{figure}[b]
    \vspace{-20pt}
    \centering
    \subfloat[Sinusoidal goal tracking]{\includegraphics[width=0.24\textwidth]{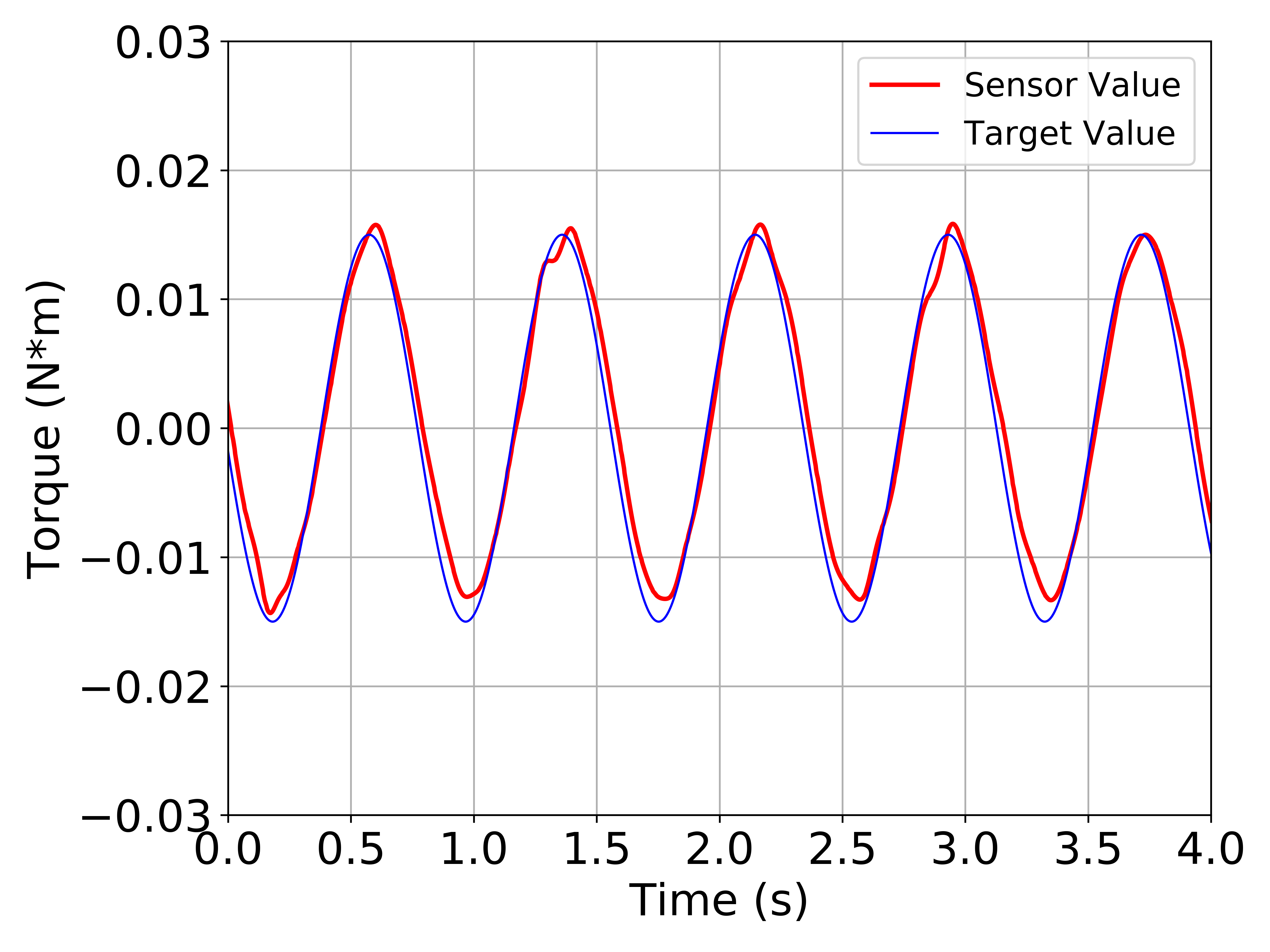}}
    \hfill
    \subfloat[Square wave goal tracking]{\includegraphics[width=0.24\textwidth]{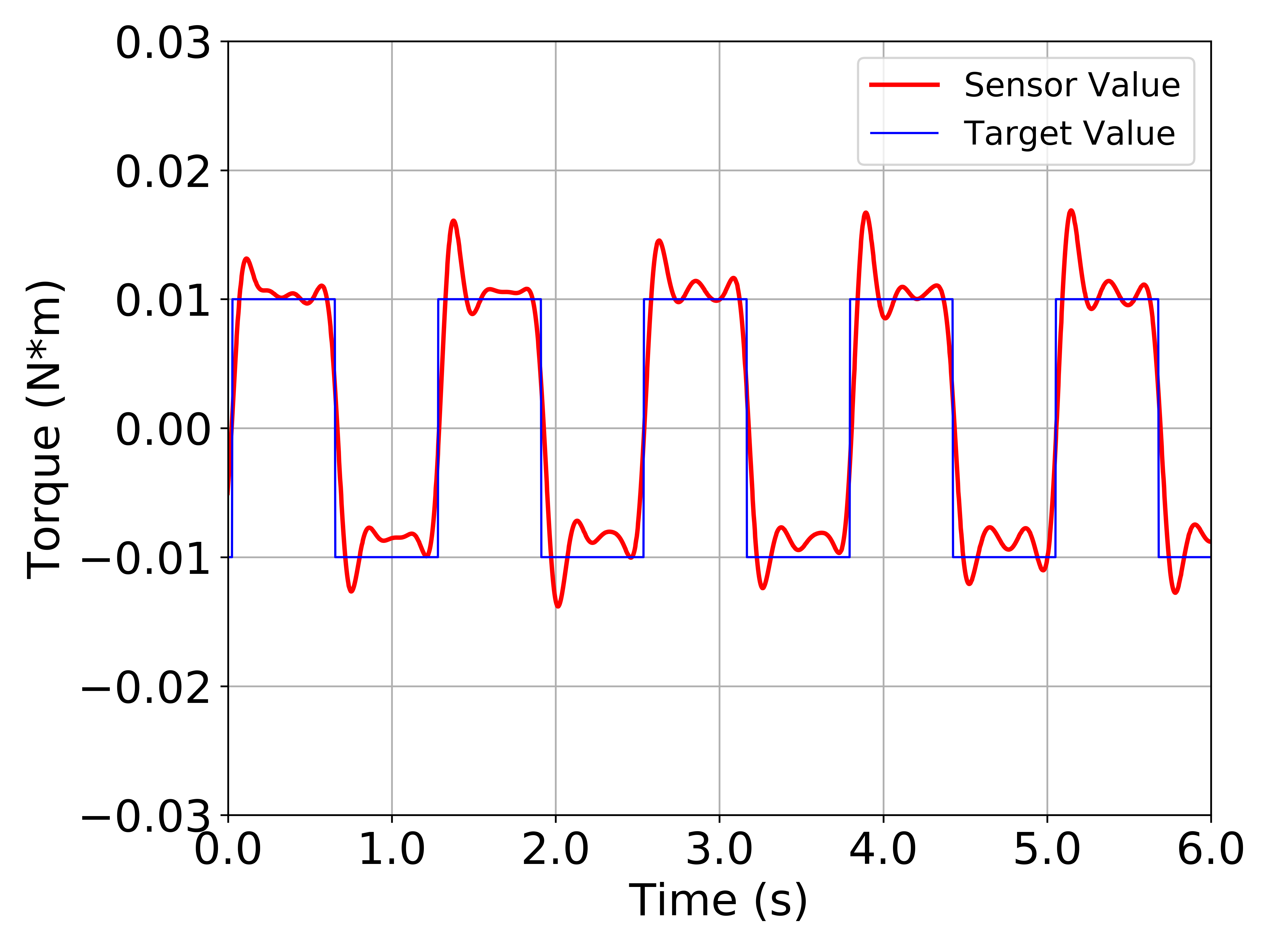}}
    \vspace{0pt}
    \caption{Tracking performance of (a) sinusoidal and (b) square Goal.}
    \label{fig:sinusoidal_square}
    \vspace{-2pt}
\end{figure}

Fig.\ref{fig:sinusoidal_square} shows torque variation curves for sinusoidal and square wave tracking. Due to sensor noise, a low-pass filter (5Hz cut-off, 4th order) was applied. The blue curve indicates the target torque, and the red curve shows the filtered response. The system accurately tracks sinusoidal goals and responds quickly to steps.

\renewcommand{\arraystretch}{0.8} 
\setlength{\aboverulesep}{2pt}  
\setlength{\belowrulesep}{3pt}  
\captionsetup{font=small}
\setlength{\tabcolsep}{4pt} 
\begin{table}[h]
    \vspace{-10pt}
    \scriptsize
    \centering
    \caption{Sinusoidal Goal Tracking Performance (5 Tests)}
    \vspace{0pt}
    \label{tab:sinusoidal_tracking}
    \begin{tabular}{lccc}
        \toprule
        $\alpha$ (N$\cdot$m) & $\omega$ & RMSE (N$\cdot$m) & $L$ (s) \\
        \midrule
        0.0150 & 8  & 0.0021 & 0.0126 \\
        0.0150 & 16 & 0.0058 & 0.0239 \\
        0.0300 & 8  & 0.0081 & 0.0253 \\
        0.0300 & 16 & 0.0095 & 0.0174 \\
        \bottomrule
    \end{tabular}
    \vspace{-15pt}
\end{table}
\renewcommand{\arraystretch}{0.8} 
\setlength{\aboverulesep}{2pt}  
\setlength{\belowrulesep}{3pt}  
\captionsetup{font=small}
\setlength{\tabcolsep}{4pt} 
\begin{table}[h]
    \vspace{-5pt}
    \scriptsize
    \centering
    \caption{Step Goal Tracking Performance (5 Tests)}
    \vspace{0pt}
    \label{tab:step_tracking_5tests}
    \begin{tabular}{lccc}
        \toprule
        $A$ (N$\cdot$m) & RMSE (N$\cdot$m) & OS (\%) & $t_p$ (s) \\
        \midrule
        0.0100 & 0.0031 & 27.38 & 0.0916 \\
        0.0200 & 0.0144 & 10.20 & 0.1800 \\
        \bottomrule
    \end{tabular}
    \vspace{-15pt}
\end{table}

\subsection{Human-Online Testing}

\renewcommand{\arraystretch}{0.8} 
\setlength{\aboverulesep}{2pt}  
\setlength{\belowrulesep}{3pt}  
\captionsetup{font=small}
\setlength{\tabcolsep}{4pt} 
\begin{table}[t]
    \vspace{5pt}
    \scriptsize
    \centering
    \caption{Real-world Testing in a Tele-played Game (5 Tests)}
    \vspace{-3pt}
    \label{tab:realworld_teleplay}
    \begin{tabular}{llcccccc}
        \toprule
        Scene & Feedback & A & B & C & D & E & Average \\
        \midrule
        \multirow{2}{*}{Discrete}  & Visual+Torque & 573.4 & 487.6 & 592.6 & 568.6 & 602.0 & 586.7 \\
        \cmidrule(lr){2-8}
                   & Visual  & 577.0 & 481.6 & 558.0 & 493.6 & 583.6 & 566.2 \\
        \midrule
        \multirow{2}{*}{Continuous} & Visual+Torque & 440.8 & 487.6 & 558.0 & 506.0 & 529.2 & 504.3 \\
        \cmidrule(lr){2-8}
                   & Visual  & 378.2 & 481.6 & 517.6 & 424.4 & 552.4 & 470.8 \\
        \bottomrule
    \end{tabular}
    \vspace{-10pt}
\end{table}

\begin{figure}[t]
    \centering
    \subfloat[Discrete Angle Test]{\includegraphics[width=0.24\textwidth]{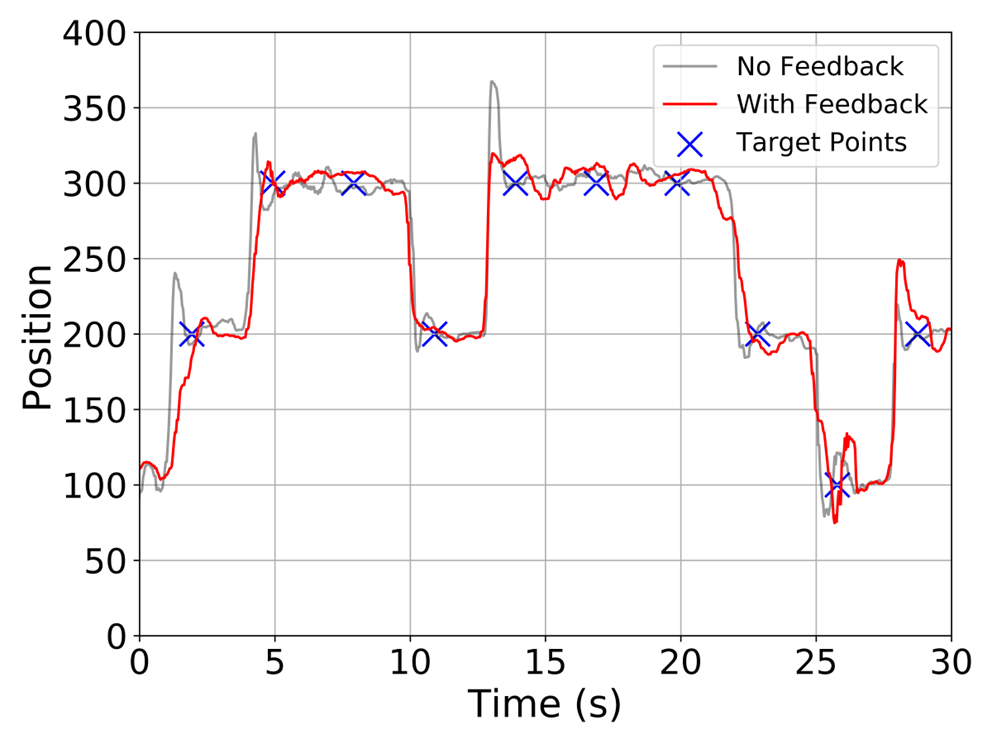}}
    \hfill
    \subfloat[Continuous Angle Test]{\includegraphics[width=0.24\textwidth]{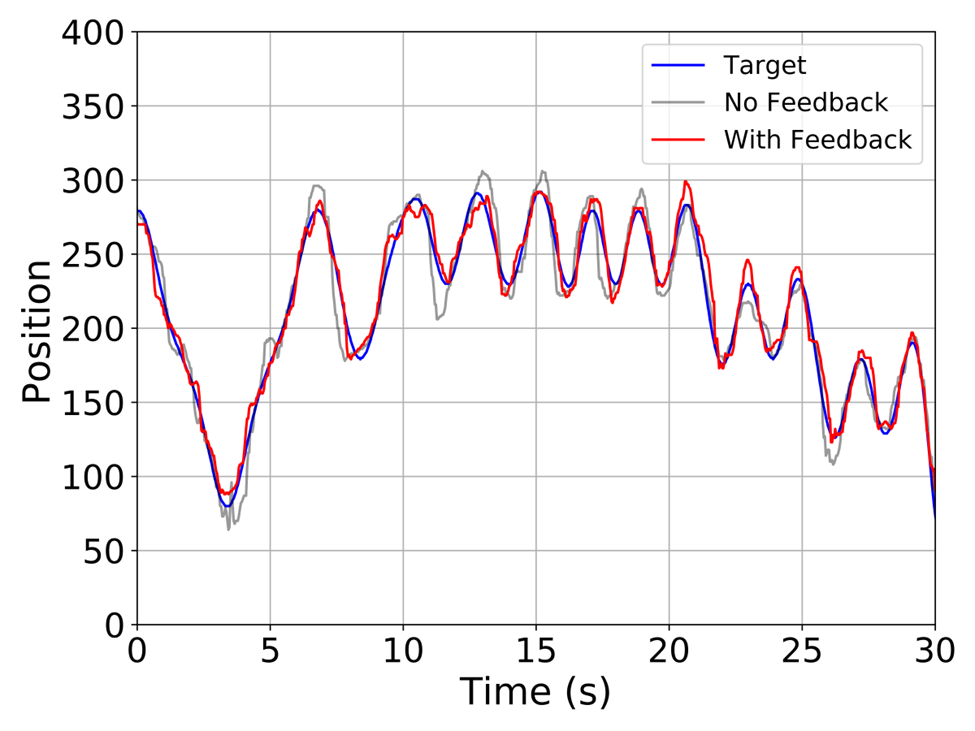}}
    \vspace{-4pt}
    \caption{Test performance of (a) discrete and (b) continuous target.}
    \label{fig:teleplay}
    \vspace{-15pt}
\end{figure}

\noindent Table~\ref{tab:realworld_teleplay} summarizes the results, where A–E denotes the five subjects. The values represent the average of five trials per condition. 
Overall, the Bi-Hap system with torque feedback outperforms the condition of the no torque feedback. In the discrete task, torque feedback improves average scores by 20.5 points, while in the continuous task, the improvement is 33.5 points. While individual gains vary, this variability likely stems from differences in tracking strategies and rotational habits. Nonetheless, the results indicate that torque feedback enhances task performance.

To further demonstrate Bi-Hap’s performance, we recorded a subject’s tracking performance in discrete and continuous angle tasks. As shown in Fig.~\ref{fig:teleplay}, torque feedback (red) improved tracking accuracy, especially during rapid transitions. The subject responded faster, corrected errors earlier, and followed the target (blue) more closely than in the no-feedback condition (gray).

In summary, the Bi-Hap system enhances in-hand dexterous telemanipulation by delivering low-latency, accurate torque feedback through a compact, momentum-based design. Experimental results demonstrate its effectiveness in improving manipulation accuracy, responsiveness, and error correction across various tasks. The integrated error-adaptive strategy further boosts operator awareness and control. In future work, we aim to extend Bi-Hap to support 3-DoF torque feedback and validate its performance on physical robotic hands for broader real-world applications.

\vspace{-3pt}
\bibliographystyle{IEEEtran} 
\bibliography{ref.bib}

\begin{thebibliography}{10}
\providecommand{\url}[1]{#1}
\csname url@samestyle\endcsname
\providecommand{\newblock}{\relax}
\providecommand{\bibinfo}[2]{#2}
\providecommand{\BIBentrySTDinterwordspacing}{\spaceskip=0pt\relax}
\providecommand{\BIBentryALTinterwordstretchfactor}{4}
\providecommand{\BIBentryALTinterwordspacing}{\spaceskip=\fontdimen2\font plus
\BIBentryALTinterwordstretchfactor\fontdimen3\font minus \fontdimen4\font\relax}
\providecommand{\BIBforeignlanguage}[2]{{%
\expandafter\ifx\csname l@#1\endcsname\relax
\typeout{** WARNING: IEEEtran.bst: No hyphenation pattern has been}%
\typeout{** loaded for the language `#1'. Using the pattern for}%
\typeout{** the default language instead.}%
\else
\language=\csname l@#1\endcsname
\fi
#2}}
\providecommand{\BIBdecl}{\relax}
\BIBdecl

\bibitem{huang2025human}
Y.~Huang, D.~Fan, D.~Yan, W.~Qi, G.~Deng, Z.~Shao, Y.~Luo, D.~Li, Z.~Wang, Q.~Liu \emph{et~al.}, ``Human-robot collaborative tele-grasping in clutter with five-fingered robotic hands,'' \emph{IEEE Robotics and Automation Letters}, 2025.

\bibitem{cortese2014mechatronic}
M.~Cortese, M.~Cempini, P.~R. de~Almeida~Ribeiro, S.~R. Soekadar, M.~C. Carrozza, and N.~Vitiello, ``A mechatronic system for robot-mediated hand telerehabilitation,'' \emph{IEEE/ASME transactions on mechatronics}, vol.~20, no.~4, pp. 1753--1764, 2014.

\bibitem{wang2024dexcap}
C.~Wang, H.~Shi, W.~Wang, R.~Zhang, L.~Fei-Fei, and C.~K. Liu, ``Dexcap: Scalable and portable mocap data collection system for dexterous manipulation,'' \emph{arXiv preprint arXiv:2403.07788}, 2024.

\bibitem{zhang2025doglove}
H.~Zhang, S.~Hu, Z.~Yuan, and H.~Xu, ``Doglove: Dexterous manipulation with a low-cost open-source haptic force feedback glove,'' \emph{arXiv preprint arXiv:2502.07730}, 2025.

\bibitem{abad2022novel}
A.~C. Abad, D.~Reid, and A.~Ranasinghe, ``A novel untethered hand wearable with fine-grained cutaneous haptic feedback,'' \emph{Sensors}, vol.~22, no.~5, p. 1924, 2022.

\bibitem{liu2014development}
L.~Liu, S.~Miyake, N.~Maruyama, K.~Akahane, and M.~Sato, ``Development of two-handed multi-finger haptic interface spidar-10,'' in \emph{Haptics: Neuroscience, Devices, Modeling, and Applications: 9th International Conference, EuroHaptics 2014, Versailles, France, June 24-26, 2014, Proceedings, Part II 9}.\hskip 1em plus 0.5em minus 0.4em\relax Springer, 2014, pp. 176--183.

\bibitem{meijneke2021design}
C.~Meijneke, B.~Sterke, G.~Hermans, W.~Gregoor, H.~Vallery, and D.~Lemus, ``Design and evaluation of pint-sized gyroscopic actuators,'' in \emph{2021 IEEE/ASME International Conference on Advanced Intelligent Mechatronics (AIM)}.\hskip 1em plus 0.5em minus 0.4em\relax IEEE, 2021, pp. 454--461.

\bibitem{berna2024hapticwhirl}
J.~L. Berna~Moya, A.~van Oosterhout, M.~T. Marshall, and D.~Martinez~Plasencia, ``Hapticwhirl, a flywheel-gimbal handheld haptic controller for exploring multimodal haptic feedback,'' \emph{Sensors}, vol.~24, no.~3, p. 935, 2024.

\bibitem{hashimoto2022metamorphx}
T.~Hashimoto, S.~Yoshida, and T.~Narumi, ``Metamorphx: An ungrounded 3-dof moment display that changes its physical properties through rotational impedance control,'' in \emph{Proceedings of the 35th Annual ACM Symposium on User Interface Software and Technology}, 2022, pp. 1--14.

\bibitem{wang2024real}
H.~Wang, H.~Bai, X.~Zhang, Y.~Jung, M.~Bowman, and L.~Tao, ``Real-time dexterous telemanipulation with an end-effect-oriented learning-based approach,'' in \emph{2024 IEEE/RSJ International Conference on Intelligent Robots and Systems (IROS)}.\hskip 1em plus 0.5em minus 0.4em\relax IEEE, 2024, pp. 12\,164--12\,169.

\bibitem{pacchierotti2023cutaneous}
C.~Pacchierotti and D.~Prattichizzo, ``Cutaneous/tactile haptic feedback in robotic teleoperation: Motivation, survey, and perspectives,'' \emph{IEEE Transactions on Robotics}, vol.~40, pp. 978--998, 2023.

\bibitem{raitor2023design}
M.~Raitor, C.~M. Nunez, P.~J. Stolka, A.~M. Okamura, and H.~Culbertson, ``Design and evaluation of haptic guidance in ultrasound-based needle-insertion procedures,'' \emph{IEEE Transactions on Biomedical Engineering}, vol.~71, no.~1, pp. 26--35, 2023.

\bibitem{quek2018evaluation}
Z.~F. Quek, W.~R. Provancher, and A.~M. Okamura, ``Evaluation of skin deformation tactile feedback for teleoperated surgical tasks,'' \emph{IEEE transactions on haptics}, vol.~12, no.~2, pp. 102--113, 2018.

\bibitem{sagardia2016platform}
M.~Sagardia, T.~Hulin, K.~Hertkorn, P.~Kremer, and S.~Sch{\"a}tzle, ``A platform for bimanual virtual assembly training with haptic feedback in large multi-object environments,'' in \emph{Proceedings of the 22nd ACM Conference on Virtual Reality Software and Technology}, 2016, pp. 153--162.

\bibitem{krieger2024motion}
K.~Krieger, Y.~De~Pra, H.~Ritter, and A.~Moringen, ``Motion analysis of upper limb and hand in a haptic rotation task,'' \emph{arXiv preprint arXiv:2411.12765}, 2024.

\bibitem{walker2017haptic}
J.~M. Walker, H.~Culbertson, M.~Raitor, and A.~M. Okamura, ``Haptic orientation guidance using two parallel double-gimbal control moment gyroscopes,'' \emph{IEEE transactions on haptics}, vol.~11, no.~2, pp. 267--278, 2017.

\bibitem{verret2024synthesis}
S.~Verret, T.~Lalibert{\'e}, R.~Cloutier, and C.~Gosselin, ``Synthesis, dynamic modeling, prototyping and control of a handheld rotational inertia generator,'' \emph{IEEE Transactions on Haptics}, vol.~17, no.~4, pp. 591--603, 2024.

\bibitem{skuric2022simplefoc}
A.~Skuric, H.~S. Bank, R.~Unger, O.~Williams, and D.~Gonz{\'a}lez-Reyes, ``Simplefoc: a field oriented control (foc) library for controlling brushless direct current (bldc) and stepper motors,'' \emph{Journal of Open Source Software}, vol.~7, no.~74, p. 4232, 2022.

\bibitem{patel2020ziegler}
V.~V. Patel, ``Ziegler-nichols tuning method: Understanding the pid controller,'' \emph{Resonance}, vol.~25, no.~10, pp. 1385--1397, 2020.

\bibitem{bai2023robotic}
N.~Bai, Y.~Xue, S.~Chen, L.~Shi, J.~Shi, Y.~Zhang, X.~Hou, Y.~Cheng, K.~Huang, W.~Wang \emph{et~al.}, ``A robotic sensory system with high spatiotemporal resolution for texture recognition,'' \emph{Nature Communications}, vol.~14, no.~1, p. 7121, 2023.

\bibitem{wang2024toward}
Y.~Wang, H.~Wu, T.~Li, J.~Wang, Z.~Wei, and H.~Wang, ``Toward human-like touch sense via a bioinspired soft finger with self-decoupled bending and force sensing,'' \emph{Cell Reports Physical Science}, vol.~5, no.~10, 2024.

\bibitem{todorov2012physics}
E.~Todorov, T.~Erez, and Y.~T. MuJoCo, ``A physics engine for model-based control,'' in \emph{Proceedings of the 2012 IEEE/RSJ International Conference on Intelligent Robots and Systems}, pp. 5026--5033.

\end{thebibliography}

\end{document}